\pdfoutput=1

\documentclass[11pt]{article}
\usepackage[preprint]{acl}
\usepackage{times}
\usepackage{latexsym}

\usepackage[T1]{fontenc}
\usepackage[utf8]{inputenc}

\usepackage{microtype}
\usepackage{inconsolata}

\usepackage{graphicx}
\usepackage{subfigure}

\newcommand{\ours}{\texttt{ERD}}
\newcommand{\extract}{\texttt{Extraction}}
\newcommand{\thought}{\texttt{Reasoning}}
\newcommand{\mad}{\texttt{Debate}}

\makeatletter
\newcommand\footnoteref[1]{\protected@xdef\@thefnmark{\ref{#1}}\@footnotemark}
\makeatother

\title{\ours{}: A Framework for Improving LLM Reasoning \\for Cognitive Distortion Classification}

\author{
    Sehee Lim\thanks{Equal contribution} \\
    Yonsei University \\
    \texttt{sehee0706@yonsei.ac.kr} \\
    \And
    Yejin Kim$^\ast$ \\
    Yonsei University \\
    \texttt{yjkim.stat@yonsei.ac.kr}\\
    \And
    Chi-Hyun Choi$^\ast$ \\
    EverEx \\
    \texttt{leo@everex.co.kr} \\
    \AND
    Jy-yong Sohn\thanks{Corresponding authors} \\ 
    Yonsei University \\
    EverEx \\
    \texttt{jysohn1108@yonsei.ac.kr} \\
    \And
    Byung-Hoon Kim$^\dagger$ \\
    Yonsei University \\
    EverEx \\
    \texttt{egyptdj@yonsei.ac.kr} \\
  }

\begin{document}
\maketitle

\begin{abstract}
Improving the accessibility of psychotherapy with the aid of Large Language Models (LLMs) is garnering a significant attention in recent years. Recognizing cognitive distortions from the interviewee's utterances can be an essential part of psychotherapy, especially for cognitive behavioral therapy. 
In this paper, we propose \ours{}, which improves LLM-based cognitive distortion classification performance with the aid of additional modules of (1) extracting the parts related to cognitive distortion, and (2) debating the reasoning steps by multiple agents. 
Our experimental results on a public dataset show that \ours{} improves the multi-class F1 score as well as binary specificity score. Regarding the latter score, it turns out that our method is effective in debiasing the baseline method which has high false positive rate, especially when the summary of multi-agent debate is provided to LLMs.
\end{abstract}

\section{Introduction}
\vspace{-2mm}
Large Language Models (LLMs) are dominating the research areas in machine learning and artificial intelligence, broadening its usage in various applications~\citep{radford2018improving,radford2019language,brown2020language,openai2023gpt,ouyang2022training}. Especially in the medical domain, PaLM~\citep{chowdhery2022palm} and its variants, such as Med-PaLM~\citep{singhal2022large}, are equipped with medical data and instructions to answer the questions from clinical field~\citep{chowdhery2022palm, singhal2023towards}.
In addition, conversational AI assistant chatbots are devised to support patients with mental health issues~\citep{rathje2023gpt, Aditya-et-a1,saha-etal-2022-shoulder, stock-et-al-2023, liu2023chatcounselor, welivita-etal-2021-large, sharma2020computational}.

Recognizing the fact that individuals with mental disorders hesitate to seek in-person medical consultations~\citep{herbert1980resistance}, previous studies~\citep{yang2023psycot,lee2023chain,chen-etal-2023-empowering} attempt to enhance the accessibility and quality of psychotherapy through the use of LLMs with Chain-of-Thought (CoT) reasoning~\citep{wei2022chain}. These models aim to detect the user's personality and interpret their mental state in order to generate more empathetic responses.

For example, Diagnosis-of-Thought (DoT) uses LLMs to classify cognitive distortions from utterances, which is a crucial part of Cognitive Behavior Therapy (CBT)~\citep{chen-etal-2023-empowering}.

While the DoT method holds promise, one key challenge that remains an open issue is the tendency of the model to overdiagnose cognitive distortions, incorrectly inferring irrational thought patterns even when the user's statements are benign. In addition, the distortion classification performance of DoT in multi-class setup is close to that of random guessing, which limits its usage in practice.

\begin{figure*}[t]
\vspace{-5mm}
\begin{center}
\includegraphics[width=\linewidth]{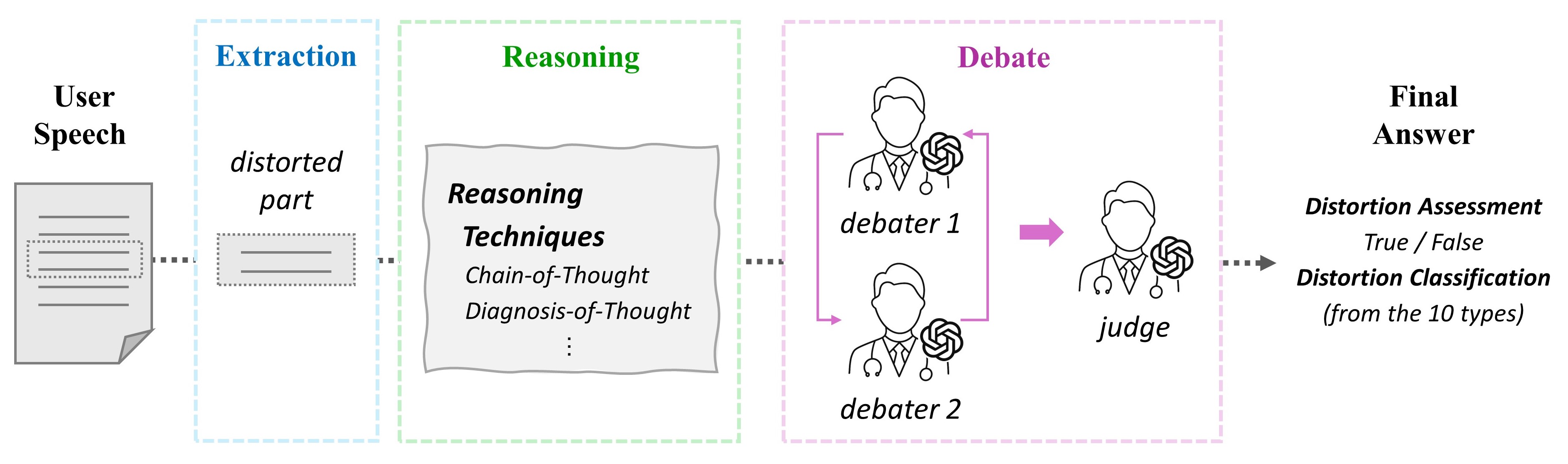}
\vspace{-8mm}
\caption{\label{fig:overview}
The pipeline of Extraction-Reasoning-Debate (\ours{}), which 
detects and classify the cognitive distortion from the input user speech. It begins with the identification and extraction of potential cognitive distortions from the user speech. These extracted elements are then utilized to construct an intermediate reasoning step. Subsequently, a debate is conducted, wherein multiple LLM agents deliberate to assess the presence and type of cognitive distortion. Finally, a judge integrates the entire debate process to get the final answer on the distortion classification problem.
}
\end{center}
\end{figure*}

In this paper, we tackle these issues by proposing a new framework for classifying cognitive distortions from the user utterances, by introducing modules for debiasing the overdiagnosing tendency of existing methods and for improving the performance on classifying distortion types inferred from the utterances. 

Our main contributions can be summarized as below:

\begin{itemize}
    \item We introduce \ours{}, a new framework for classifying cognitive distortions in the user utterances using three steps: \extract{}, \thought{}, and \mad{}, each of which uses LLMs.  The first step lets LLM extract a part of the utterances that is related with the distortion, the second step uses LLM to generate the thought process of estimating cognitive distortions from the extracted part, and the third step uses multi-agent LLMs to discuss the thought process described in the second step and make the final decision. 
    \item Compared with existing baselines, \ours{} improves the multi-class F1 score for distortion classification task by more than 9\% and improves the distortion assessment specificity score by more than 25\%, when tested on the cognitive distortion detection dataset with 2530 samples in Kaggle. 
    \item We provide factor analysis on \ours{}, showing that (1) multiple rounds of debate in \ours{} is beneficial for improving the classification score, and (2) the summarization and the validity evaluation processes during the debate step enhance the debiasing effect.
\end{itemize}

\section{\ours{}}
\vspace{-2mm}

We propose Extraction-Reasoning-Debate (\ours{}), a framework for classifying distortions in a given user speech, as shown in Fig.~\ref{fig:overview}. Below we elaborate each step in our framework.

\subsection{Extraction}
\vspace{-2mm}

\begin{table}[b!]
\centering
\small
\setlength{\tabcolsep}{3pt}
\begin{tabular}{l|c}
\hline
\textbf{Input}         & \textbf{Distortion Classification}  \\ \hline
User Speech    & $15.28_{0.65}$ \\
Distorted Part of User Speech & $\textbf{27.08}_{0.27}$ \\\hline
\end{tabular}
\vspace{-2mm}
\caption{
\label{tab:gt_ext}
Multi-class F1 score of
DoT~\citep{chen-etal-2023-empowering}
for the cognitive distortion classification problem, when two different inputs are given. The first option uses the user speech as the input, as done in~\citep{chen-etal-2023-empowering}. The second option is considered by us, which only puts the ground-truth distorted part within the user speech.
Putting only the distorted part significantly improves the classification performance, which motivates the \texttt{Extraction} step in \ours{} framework.}
\end{table}

To provide the motivation for the \extract{} step proposed in our method, we first share our empirical results showing that 
extracting the distorted parts of user speech is beneficial for distortion classification. 
Table~\ref{tab:gt_ext} shows the multi-class F1 score of Diagnosis-of-Thought (DoT) method for distortion classification problem (predicting out of 10 classes), tested on a cognitive distortion detection dataset with 2530 samples in Kaggle\footnote{\label{dataset}\url{https://www.kaggle.com/datasets/sagarikashreevastava/cognitive-distortion-detetction-dataset}}. We test on two different options: (1) putting the user speech as it is, and (2) putting the ground-truth part (provided in the `distorted part' column of the dataset) within the speech, that indicates the distortion. Table~\ref{tab:gt_ext} shows that the multi-class F1 score increases more than 10\% when the ground-truth distorted part is extracted before running DoT. 

Motivated by this result, prior to the \thought{} step (e.g., DoT) which outputs the thought process for assessing/classifying the distortion, we add an \extract{} step which instructs LLMs isolate the segments from the user's utterance that may potentially exhibit cognitive distortions. This process of extraction is done \emph{without} paraphrasing or summarizing, thereby preserving the original context and nuances for the subsequent thought process. In summary, \extract{} process ensures that the LLMs' responses hinge on the most informative facets of the utterance, which in turn enhance the quality of the distortion classification performance.

\begin{table*}
\centering
\small
\begin{tabular}{l|ccc|c}
\hline
&  \multicolumn{3}{|c|}{\textbf{Distortion Assessment} (True/False)}   & \textbf{Distortion Classification} (out of 10 types) \\ 
\textbf{Method}&Sensitivity  & Specificity&F1 Score   & Weighted F1 Score \\
\hline

\thought{}          &$\underline{99.29}_{ 0.19}$&$ 6.79_{ 0.34}$ &$ \textbf{78.26}_{ 0.16}$&$ 15.28_{ 0.65}$\\
+\extract{}       &$ \textbf{99.83}_{ 0.03}$   &$ 0.93_{ 0.22}$             &$\underline{77.48}_{ 0.04}$&$ \textbf{24.40}_{ 0.69}$\\
+\mad{}           &$ 73.10_{ 0.26}$            &$ \textbf{33.05}_{ 0.58}$   &$ 68.89_{ 0.24}$           &$ 22.18_{ 0.99}$\\
+\extract{}+\mad{}&$ 74.89_{ 2.31}$            &$ \underline{30.74}_{ 3.92}$&$ 69.49_{ 0.62}$           &$ \underline{24.27}_{ 1.14}$\\ \hline
\end{tabular}
\vspace{-2mm}
\caption{\label{tab:ablation}
Cognitive distortion assessment/classification results of \ours{} when various modules (\extract{} and \mad{}) are added. Here, we test on cognitive distortion detection dataset in Kaggle, and 
use DoT~\citep{chen-etal-2023-empowering} method for the \thought{} step. Upon the above results, \extract{} improves the distortion classification performance and \mad{} increases the distortion assessment specificity significantly. Combining both \extract{} and \mad{} takes the sweet spot, simultaneously enhancing both performances.}
\end{table*}

\begin{figure}[t!]
\vskip 0.2in
\vspace{-5mm}
\begin{center}
\includegraphics[width=0.9\columnwidth]{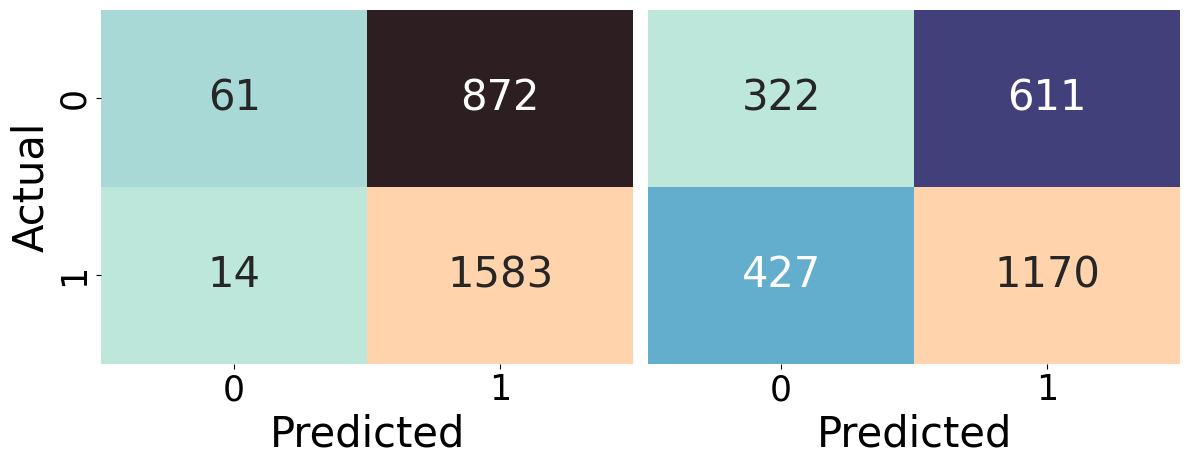}
\vspace{-3mm}
\caption{Confusion matrices of \ours{} when tested on 2530 samples: (Left) only \thought{} is used, (Right) \extract{}, \thought{} and \mad{} steps are used. Including \extract{} and \mad{} modules increases the number of true negatives from 61 to 322, thus correctly identifying the samples with `no distortion'.
}
\label{fig:combined_confusion_matrix}
\end{center}
\vskip -0.2in
\end{figure}

\subsection{Reasoning} 
\vspace{-2mm}
Our target task (cognitive distortion classification from the user speech) is naturally considered as a task that requires logical thinking, if we imagine how doctors classify the patients. In recent years, various methods propose letting LLMs mimic the logical thought process or reasoning steps. For example, chain-of-thought (CoT) prompting and its variants~\citep{wei2022chain,kojima2022large,yao2023tree,besta2023graph, chen2023program, yang2023psycot, lee2023chain} provide a significant performance improvement in various reasoning tasks including common sense reasoning and mathematical reasoning. 

Our \thought{} step chooses any existing methods which let LLMs output the thought process for performing the target task. By default, we use diagnosis-of-thought (DoT)~\citep{chen-etal-2023-empowering}
comprised of three critical stages (subjectivity assessment, contrastive reasoning, and schema analysis) that construct rationales for the detection of cognitive distortions. 
At the \emph{subjectivity assessment} stage, the input utterances are differentiated between the objective facts and the subjective thoughts. 
This is followed by the \emph{contrastive reasoning} stage, where the process elicits both supportive and contradictory perspectives to the speaker's viewpoint.
The final stage, \emph{schema analysis}, involves delving into the underlying thought schema. 

\subsection{Debate}
\vspace{-2mm}

Several recent works on using LLMs for reasoning tasks show that multiple LLM agents debating their thought processes significantly improve the performance~\citep{liang2023encouraging, zheng2023judging, xiong2023examining, chan2023chateval, du2023improving}.
Motivated by this observation, we add multi-agent debate (or \mad{}) step following the \thought{} step. 
In Figure~\ref{fig:overview}, \ours{} employs three LLM agents, each designated with the role of ``physician'' to simulate a professional medical debate. The discussion between first two agents (two debators) is overseen by the third agent (called the judge agent), bearing the role of ``head doctor'' who monitors the entire debate to ensure a fair evaluation. The third agent is introduced, motivated by recent result showing that LLMs can behave as a good judge~\citep{zheng2023judging}.
The first debater presents arguments for the presence or absence of cognitive distortion in the user speech, based on the LLM outputs obtained in the \extract{} and \thought{} steps. 
Subsequently, the second debator counters the initial assertions, presenting a contradicting viewpoint. 
The first debater then responds to this counterargument, followed by a second round of rebuttal from the second debater, resulting in two rounds of argumentation.
One can consider repeating this iterative exchange of thoughts for multiple rounds. 
After this iterative process, the judge agent integrates the entire discourse, employing two proposed methodologies to reach a final decision. 

We consider two different options for controlling the behavior of the judge agent to get better performances. 
The first option involves a straightforward summarization of the total debate process. 
The second option involves summarizing the debate and evaluating which side's arguments are more valid.
By adding such summarization process, we expect that the final answer of \ours{} is based on a comprehensive consideration of all presented viewpoints.

\begin{table*}
\centering
\small
\begin{tabular}{l|ccc|c}
\hline
& \multicolumn{3}{|c|}{\textbf{Distortion Assessment}}  & \textbf{Distortion Classification} \\ 
& Sensitivity   & Specificity   & F1 Score  & Weighted F1 Score \\
\hline

\ours{} without Summarization &$\textbf{92.13}_{ 0.38}$&$11.01_{ 0.66}$&$ \textbf{75.48}_{ 0.23}$&$ \textbf{25.28}_{ 0.46}$\\
+Summarization    &$ \underline{86.10}_{ 0.58}$&$ \underline{19.58}_{ 1.36}$&$ \underline{73.88}_{ 0.36}$&$ 23.96_{ 1.05}$\\
+Summarization+Validity Evaluation       &$74.89_{ 2.31}$ & $\textbf{30.74}_{ 3.92}$&$ 69.49_{ 0.62}$ & $\underline{24.27}_{ 1.14}$\\
\hline
\end{tabular}
\vspace{-2mm}
\caption{\label{tab:summary}
Comparison of \ours{} with three different prompting options that control the behavior of the judge.
For the first option, judge predicts the cognitive distortion type only based on the debate process log, without any summarization step. For the second option, judge first \emph{summarizes} the debate and then predicts the cognitive distortion type. In the final option, judge \emph{summarizes} the debate, \emph{evaluates the validity} of the claims in the debate, and then predicts the cognitive distortion type.
Both summarization and validity evaluation steps improve the performance in terms of specificity. 
}
\end{table*}
\section{Experiments}
\paragraph{Settings}
\vspace{-2mm}
We use a cognitive distortion detection dataset~\citep{shreevastava-foltz-2021-detecting} composed of speeches that correspond to 10 types of cognitive distortions and neutral speeches. 
This dataset, sourced from Kaggle\footnoteref{dataset}, contains 2530 annotated examples by experts and is designed to facilitate two tasks: distortion assessment and distortion classification. 
In the distortion assessment task, the model determines whether cognitive distortion is present in the patient's utterance. 
In the distortion classification task, the model identifies the specific type of cognitive distortion.
We report the Sensitivity, Specificity and F1 score for the distortion assessment task, and the weighted F1 score for the distortion classification task.
We run 3 random trials and report the mean and standard deviation values. 
We employ the model gpt-3.5-turbo, and set the temperature to 0.1.

\paragraph{Experimental results}

Table~\ref{tab:ablation} shows the performances of \ours{}, when different modules are plugged in. Compared with naive method using \thought{} module only, adding \extract{} module improves the distortion classification score by more than 9\%, and adding \mad{} module not only improves the distortion classification score by around 7\%, but also improves the distortion assessment specificity by more than 25\%. 
Fig.~\ref{fig:combined_confusion_matrix} shows the confusion matrix of the \ours{} for two cases: (1) when only the \thought{} module is used, and (2) when \extract{}, \thought{} and \mad{} are used. This result shows that adding \extract{} and \mad{} modules promotes the correct estimation of utterances with no distortion.

Recall that in \mad{} step of \ours{}, we consider different prompting techniques to control the behavior of the judge agent when making the final decision. Table~\ref{tab:summary} shows the effect of such prompts for three variants: 

(1) ``\ours{} without summarization'' does not instruct judge to summarize the claims of debate and just directly make decision, (2) ``\ours{} with summarization'' instructs judge to summarize the claims before making the decision, and ``\ours{} with summarization and validity evaluation'' instructs judge to summarize and evaluate the claims of debate before making the decision.
Note that the specificity is keep improved as we provide more detailed instructions to the judge agent.

\begin{table}[t!]
\centering
\scriptsize
\begin{tabular}{l|ccc}
\hline
\textbf{Metric}          & \textbf{Round 1} & \textbf{Round 2} & \textbf{Round 3} \\ \hline
Binary F1 & $52.13_{1.25}$ & $69.49_{0.62}$  & $\textbf{70.74}_{0.44}$\\
Multi-class F1    & $22.79_{1.62}$ & $24.27_{1.14}$ & $\textbf{24.83}_{0.81}$ \\ \hline
\end{tabular}
\vspace{-2mm}
\caption{\label{tab:by_round}
F1 scores for different $r$, the number of \mad{} rounds. 
The performances improve as $r$ increases. 
}
\end{table}

Table~\ref{tab:by_round} shows how the performance improves as we increase $r$, the number of \mad{} rounds used in \ours{}. The results show that increasing the number of \mad{} rounds led to enhancements in both the binary F1 score and the multi-class F1 score. The performance saturates after $r=2$, thus better to use two rounds of debate considering the token efficiency. 
This finding aligns with the results presented in a related work on multi-agent debate of LLMs, demonstrating a similar pattern in the impact of the number of debate rounds on the model performance~\citep{du2023improving}.

\section{Conclusion}
\vspace{-2mm}
We introduce \ours{}, a framework 
using LLMs to estimate the cognitive distortion contained in the user utterances through three steps: \texttt{Extracting} distorted parts within the utterances, \texttt{Reasoning} the estimation of the corresponding distortion classes, and \texttt{Debating} the initial estimation using multiple agents.   
Compared with existing baselines only having the \texttt{reasoning} step, including the \texttt{extraction} and \texttt{debating} steps improve the distortion classification performance by 9\% and improve the distortion assessment specificity by over 25\%. 
Such improvements is crucial to cognitive behavior therapy since \ours{} is more adept at correctly identifying cases without distortions, avoiding the pitfall of over-diagnosing cognitive distortions.
Furthermore, experimental results reveal that we can control the behavior of \ours{} with various prompting options.

% \bibliography{custom}

\newpage


\begin{thebibliography}{30}
  \expandafter\ifx\csname natexlab\endcsname\relax\def\natexlab#1{#1}\fi
  
  \bibitem[{Besta et~al.(2023)Besta, Blach, Kubicek, Gerstenberger, Gianinazzi, Gajda, Lehmann, Podstawski, Niewiadomski, Nyczyk et~al.}]{besta2023graph}
  Maciej Besta, Nils Blach, Ales Kubicek, Robert Gerstenberger, Lukas Gianinazzi, Joanna Gajda, Tomasz Lehmann, Michal Podstawski, Hubert Niewiadomski, Piotr Nyczyk, et~al. 2023.
  \newblock Graph of thoughts: Solving elaborate problems with large language models.
  \newblock \emph{arXiv preprint arXiv:2308.09687}.
  
  \bibitem[{Brown et~al.(2020)Brown, Mann, Ryder, Subbiah, Kaplan, Dhariwal, Neelakantan, Shyam, Sastry, Askell et~al.}]{brown2020language}
  Tom Brown, Benjamin Mann, Nick Ryder, Melanie Subbiah, Jared~D Kaplan, Prafulla Dhariwal, Arvind Neelakantan, Pranav Shyam, Girish Sastry, Amanda Askell, et~al. 2020.
  \newblock Language models are few-shot learners.
  \newblock \emph{Advances in neural information processing systems}, 33:1877--1901.
  
  \bibitem[{Chan et~al.(2023)Chan, Chen, Su, Yu, Xue, Zhang, Fu, and Liu}]{chan2023chateval}
  Chi-Min Chan, Weize Chen, Yusheng Su, Jianxuan Yu, Wei Xue, Shanghang Zhang, Jie Fu, and Zhiyuan Liu. 2023.
  \newblock \href {http://arxiv.org/abs/2308.07201} {Chateval: Towards better llm-based evaluators through multi-agent debate}.
  
  \bibitem[{Chen et~al.(2023{\natexlab{a}})Chen, Ma, Wang, and Cohen}]{chen2023program}
  Wenhu Chen, Xueguang Ma, Xinyi Wang, and William~W. Cohen. 2023{\natexlab{a}}.
  \newblock \href {http://arxiv.org/abs/2211.12588} {Program of thoughts prompting: Disentangling computation from reasoning for numerical reasoning tasks}.
  
  \bibitem[{Chen et~al.(2023{\natexlab{b}})Chen, Lu, and Wang}]{chen-etal-2023-empowering}
  Zhiyu Chen, Yujie Lu, and William Wang. 2023{\natexlab{b}}.
  \newblock \href {https://doi.org/10.18653/v1/2023.findings-emnlp.284} {Empowering psychotherapy with large language models: Cognitive distortion detection through diagnosis of thought prompting}.
  \newblock In \emph{Findings of the Association for Computational Linguistics: EMNLP 2023}, pages 4295--4304, Singapore. Association for Computational Linguistics.
  
  \bibitem[{Chowdhery et~al.(2022)Chowdhery, Narang, Devlin, Bosma, Mishra, Roberts, Barham, Chung, Sutton, Gehrmann et~al.}]{chowdhery2022palm}
  Aakanksha Chowdhery, Sharan Narang, Jacob Devlin, Maarten Bosma, Gaurav Mishra, Adam Roberts, Paul Barham, Hyung~Won Chung, Charles Sutton, Sebastian Gehrmann, et~al. 2022.
  \newblock Palm: Scaling language modeling with pathways.
  \newblock \emph{arXiv preprint arXiv:2204.02311}.
  
  \bibitem[{Du et~al.(2023)Du, Li, Torralba, Tenenbaum, and Mordatch}]{du2023improving}
  Yilun Du, Shuang Li, Antonio Torralba, Joshua~B. Tenenbaum, and Igor Mordatch. 2023.
  \newblock \href {http://arxiv.org/abs/2305.14325} {Improving factuality and reasoning in language models through multiagent debate}.
  
  \bibitem[{Kojima et~al.(2022)Kojima, Gu, Reid, Matsuo, and Iwasawa}]{kojima2022large}
  Takeshi Kojima, Shixiang~Shane Gu, Machel Reid, Yutaka Matsuo, and Yusuke Iwasawa. 2022.
  \newblock Large language models are zero-shot reasoners.
  \newblock \emph{Advances in neural information processing systems}, 35:22199--22213.
  
  \bibitem[{Lee et~al.(2023)Lee, Lee, Shin, Bae, and Hahn}]{lee2023chain}
  Yoon~Kyung Lee, Inju Lee, Minjung Shin, Seoyeon Bae, and Sowon Hahn. 2023.
  \newblock Chain of empathy: Enhancing empathetic response of large language models based on psychotherapy models.
  \newblock \emph{arXiv preprint arXiv:2311.04915}.
  
  \bibitem[{Liang et~al.(2023)Liang, He, Jiao, Wang, Wang, Wang, Yang, Tu, and Shi}]{liang2023encouraging}
  Tian Liang, Zhiwei He, Wenxiang Jiao, Xing Wang, Yan Wang, Rui Wang, Yujiu Yang, Zhaopeng Tu, and Shuming Shi. 2023.
  \newblock \href {http://arxiv.org/abs/2305.19118} {Encouraging divergent thinking in large language models through multi-agent debate}.
  
  \bibitem[{Liu et~al.(2023)Liu, Li, Cao, Ren, Liao, and Wu}]{liu2023chatcounselor}
  June~M Liu, Donghao Li, He~Cao, Tianhe Ren, Zeyi Liao, and Jiamin Wu. 2023.
  \newblock Chatcounselor: A large language models for mental health support.
  \newblock \emph{arXiv preprint arXiv:2309.15461}.
  
  \bibitem[{OpenAI(2023)}]{openai2023gpt}
  R~OpenAI. 2023.
  \newblock Gpt-4 technical report.
  \newblock \emph{arXiv}, pages 2303--08774.
  
  \bibitem[{Ouyang et~al.(2022)Ouyang, Wu, Jiang, Almeida, Wainwright, Mishkin, Zhang, Agarwal, Slama, Ray et~al.}]{ouyang2022training}
  Long Ouyang, Jeffrey Wu, Xu~Jiang, Diogo Almeida, Carroll Wainwright, Pamela Mishkin, Chong Zhang, Sandhini Agarwal, Katarina Slama, Alex Ray, et~al. 2022.
  \newblock Training language models to follow instructions with human feedback.
  \newblock \emph{Advances in Neural Information Processing Systems}, 35:27730--27744.
  
  \bibitem[{Radford et~al.(2018)Radford, Narasimhan, Salimans, Sutskever et~al.}]{radford2018improving}
  Alec Radford, Karthik Narasimhan, Tim Salimans, Ilya Sutskever, et~al. 2018.
  \newblock Improving language understanding by generative pre-training.
  
  \bibitem[{Radford et~al.(2019)Radford, Wu, Child, Luan, Amodei, Sutskever et~al.}]{radford2019language}
  Alec Radford, Jeffrey Wu, Rewon Child, David Luan, Dario Amodei, Ilya Sutskever, et~al. 2019.
  \newblock Language models are unsupervised multitask learners.
  \newblock \emph{OpenAI blog}, 1(8):9.
  
  \bibitem[{Rathje et~al.(2023)Rathje, Mirea, Sucholutsky, Marjieh, Robertson, and Van~Bavel}]{rathje2023gpt}
  Steve Rathje, Dan-Mircea Mirea, Ilia Sucholutsky, Raja Marjieh, Claire Robertson, and Jay~J Van~Bavel. 2023.
  \newblock Gpt is an effective tool for multilingual psychological text analysis.
  
  \bibitem[{Saha et~al.(2022)Saha, Reddy, Das, Saha, and Bhattacharyya}]{saha-etal-2022-shoulder}
  Tulika Saha, Saichethan Reddy, Anindya Das, Sriparna Saha, and Pushpak Bhattacharyya. 2022.
  \newblock \href {https://doi.org/10.18653/v1/2022.naacl-main.174} {A shoulder to cry on: Towards a motivational virtual assistant for assuaging mental agony}.
  \newblock In \emph{Proceedings of the 2022 Conference of the North American Chapter of the Association for Computational Linguistics: Human Language Technologies}, pages 2436--2449, Seattle, United States. Association for Computational Linguistics.
  
  \bibitem[{Sharma et~al.(2020)Sharma, Miner, Atkins, and Althoff}]{sharma2020computational}
  Ashish Sharma, Adam~S. Miner, David~C. Atkins, and Tim Althoff. 2020.
  \newblock \href {http://arxiv.org/abs/2009.08441} {A computational approach to understanding empathy expressed in text-based mental health support}.
  
  \bibitem[{Shreevastava and Foltz(2021)}]{shreevastava-foltz-2021-detecting}
  Sagarika Shreevastava and Peter Foltz. 2021.
  \newblock \href {https://doi.org/10.18653/v1/2021.clpsych-1.17} {Detecting cognitive distortions from patient-therapist interactions}.
  \newblock In \emph{Proceedings of the Seventh Workshop on Computational Linguistics and Clinical Psychology: Improving Access}, pages 151--158, Online. Association for Computational Linguistics.
  
  \bibitem[{Singhal et~al.(2022)Singhal, Azizi, Tu, Mahdavi, Wei, Chung, Scales, Tanwani, Cole-Lewis, Pfohl et~al.}]{singhal2022large}
  Karan Singhal, Shekoofeh Azizi, Tao Tu, S~Sara Mahdavi, Jason Wei, Hyung~Won Chung, Nathan Scales, Ajay Tanwani, Heather Cole-Lewis, Stephen Pfohl, et~al. 2022.
  \newblock Large language models encode clinical knowledge.
  \newblock \emph{arXiv preprint arXiv:2212.13138}.
  
  \bibitem[{Singhal et~al.(2023)Singhal, Tu, Gottweis, Sayres, Wulczyn, Hou, Clark, Pfohl, Cole-Lewis, Neal et~al.}]{singhal2023towards}
  Karan Singhal, Tao Tu, Juraj Gottweis, Rory Sayres, Ellery Wulczyn, Le~Hou, Kevin Clark, Stephen Pfohl, Heather Cole-Lewis, Darlene Neal, et~al. 2023.
  \newblock Towards expert-level medical question answering with large language models.
  \newblock \emph{arXiv preprint arXiv:2305.09617}.
  
  \bibitem[{Steinberg et~al.(1980)Steinberg, Torem, and Saravay}]{herbert1980resistance}
  Herbert Steinberg, Moshe Torem, and Stephen~M. Saravay. 1980.
  \newblock \href {https://doi.org/10.1001/archpsyc.1980.01780220045004} {{An Analysis of Physician Resistance to Psychiatric Consultations}}.
  \newblock \emph{Archives of General Psychiatry}, 37(9):1007--1012.
  
  \bibitem[{Stock et~al.(2023)Stock, Schl\"{o}gl, and Groth}]{stock-et-al-2023}
  Anna Stock, Stephan Schl\"{o}gl, and Aleksander Groth. 2023.
  \newblock \href {https://doi.org/10.1007/978-3-031-35894-4_13} {Tell me, what are you most afraid of? exploring the effects of agent representation on information disclosure in human-chatbot interaction}.
  \newblock In \emph{Artificial Intelligence in HCI: 4th International Conference, AI-HCI 2023, Held as Part of the 25th HCI International Conference, HCII 2023, Copenhagen, Denmark, July 23–28, 2023, Proceedings, Part II}, page 179–191, Berlin, Heidelberg. Springer-Verlag.
  
  \bibitem[{Vaidyam et~al.(2019)Vaidyam, Wisniewski, Halamka, Kashavan, and Torous}]{Aditya-et-a1}
  Aditya~Nrusimha Vaidyam, Hannah Wisniewski, John~David Halamka, Matcheri~S. Kashavan, and John~Blake Torous. 2019.
  \newblock \href {https://doi.org/10.1177/0706743719828977} {Chatbots and conversational agents in mental health: A review of the psychiatric landscape}.
  \newblock \emph{The Canadian Journal of Psychiatry}, 64(7):456--464.
  \newblock PMID: 30897957.
  
  \bibitem[{Wei et~al.(2022)Wei, Wang, Schuurmans, Bosma, Xia, Chi, Le, Zhou et~al.}]{wei2022chain}
  Jason Wei, Xuezhi Wang, Dale Schuurmans, Maarten Bosma, Fei Xia, Ed~Chi, Quoc~V Le, Denny Zhou, et~al. 2022.
  \newblock Chain-of-thought prompting elicits reasoning in large language models.
  \newblock \emph{Advances in Neural Information Processing Systems}, 35:24824--24837.
  
  \bibitem[{Welivita et~al.(2021)Welivita, Xie, and Pu}]{welivita-etal-2021-large}
  Anuradha Welivita, Yubo Xie, and Pearl Pu. 2021.
  \newblock \href {https://doi.org/10.18653/v1/2021.emnlp-main.96} {A large-scale dataset for empathetic response generation}.
  \newblock In \emph{Proceedings of the 2021 Conference on Empirical Methods in Natural Language Processing}, pages 1251--1264, Online and Punta Cana, Dominican Republic. Association for Computational Linguistics.
  
  \bibitem[{Xiong et~al.(2023)Xiong, Ding, Cao, Liu, and Qin}]{xiong2023examining}
  Kai Xiong, Xiao Ding, Yixin Cao, Ting Liu, and Bing Qin. 2023.
  \newblock \href {http://arxiv.org/abs/2305.11595} {Examining inter-consistency of large language models collaboration: An in-depth analysis via debate}.
  
  \bibitem[{Yang et~al.(2023)Yang, Shi, Wan, Quan, Wang, Wu, and Wu}]{yang2023psycot}
  Tao Yang, Tianyuan Shi, Fanqi Wan, Xiaojun Quan, Qifan Wang, Bingzhe Wu, and Jiaxiang Wu. 2023.
  \newblock \href {https://openreview.net/forum?id=pW6xXXnCQu} {Psycot: Psychological questionnaire as powerful chain-of-thought for personality detection}.
  \newblock In \emph{The 2023 Conference on Empirical Methods in Natural Language Processing}.
  
  \bibitem[{Yao et~al.(2023)Yao, Yu, Zhao, Shafran, Griffiths, Cao, and Narasimhan}]{yao2023tree}
  Shunyu Yao, Dian Yu, Jeffrey Zhao, Izhak Shafran, Thomas~L Griffiths, Yuan Cao, and Karthik Narasimhan. 2023.
  \newblock Tree of thoughts: Deliberate problem solving with large language models.
  \newblock \emph{arXiv preprint arXiv:2305.10601}.
  
  \bibitem[{Zheng et~al.(2023)Zheng, Chiang, Sheng, Zhuang, Wu, Zhuang, Lin, Li, Li, Xing, Zhang, Gonzalez, and Stoica}]{zheng2023judging}
  Lianmin Zheng, Wei-Lin Chiang, Ying Sheng, Siyuan Zhuang, Zhanghao Wu, Yonghao Zhuang, Zi~Lin, Zhuohan Li, Dacheng Li, Eric~P. Xing, Hao Zhang, Joseph~E. Gonzalez, and Ion Stoica. 2023.
  \newblock \href {http://arxiv.org/abs/2306.05685} {Judging llm-as-a-judge with mt-bench and chatbot arena}.
  
  \end{thebibliography}
\end{document}